# Bridging Saliency Detection to Weakly Supervised Object Detection Based on Self-paced Curriculum Learning

Dingwen Zhang[1], Deyu Meng[2], Long Zhao[1] and Junwei Han[1*]
[1]School of Automation, Northwestern Polytechnical University
[2]School of Mathematics and Statistics, Xi'an Jiaotong University
{zhangdingwen2006yyy, zhaolongsym, junweihan2010}@gmail.com, dymeng@mail.xjtu.edu.cn

## Abstract

Weakly-supervised object detection (WOD) is a challenging problems in computer vision. The key problem is to simultaneously infer the exact object locations in the training images and train the object detectors, given only the training images with weak image-level labels. Intuitively, by simulating the selective attention mechanism of human visual system, saliency detection technique can select attractive objects in scenes and thus is a potential way to provide useful priors for WOD. However, the way to adopt saliency detection in WOD is not trivial since the detected saliency region might be possibly highly ambiguous in complex cases. To this end, this paper first comprehensively analyzes the challenges in applying saliency detection to WOD. Then, we make one of the earliest efforts to bridge saliency detection to WOD via the self-paced curriculum learning, which can guide the learning procedure to gradually achieve faithful knowledge of multi-class objects from easy to hard. The experimental results demonstrate that the proposed approach can successfully bridge saliency detection and WOD tasks and achieve the state-of-the-art object detection results under the weak supervision.

## 1 Introduction

Object detection is one of the most fundamental yetchallenging problems in computer vision community. The most recent breakthrough was achieved by Girshick *et al.* [Girshick *et al.*, 2014], who trained the Convolutional Neural Network (CNN) by using large amount of human labelled bounding boxes to learn the powerful feature representations and object classifiers. Despite their success, the problem of object detection is still under-addressed in practice due to the heavy burden of labeling the training samples. Essentially, in this big data era, humans more desire intelligent machines which are capable of automatically discovering the intrinsic patterns from the cheaply and massively collected weakly labeled images. Thus weakly supervised object detection (WOD) systems have been gaining more interests recently.

The key problem in WOD is how to extract the exact object localizations and train the corresponding object detectors from the weakly labelled training images. In such chicken-egg problem, most methods (including the proposed one) usually use the alternative learning strategy that first provides some coarse estimation to initialize the potential object locations and then gradually train the object detectors and update object locations jointly. In this paper, we leverage saliency detection to initialize the potential object locations due to the following reasons: 1) Saliency detection [Han *et al.*, 2015; Van Nguyen and Sepulveda, 2015, Han *et al.*, 2016] aims at simulating the selective attention mechanism of human visual system to automatically select sub-regions (usually the regions containing objects of interest) in image scenes. Thus, it can be readily utilized to provide useful priors to estimate the potential object localizations and fit well to the investigated task. 2) Some recent saliency detection methods such as [Cheng *et al.*, 2015] and [Van Nguyen and Sepulveda, 2015] can process much faster than the priors, e.g., intra-class similarity [Siva and Xiang, 2011], inter-class variance [Siva *et al.*, 2012], and distance mapping relation [Shi *et al.*, 2012], adopted in the existing WOD systems. 3) Several existing works, e.g., [Siva *et al.*, 2013], have attempted to apply saliency detection techniques to WOD. However they still have not sufficiently explore the intrinsic bridge between these two tasks, which motivates us to clarify the insightful relationship between these two tasks and further develop powerful learning regime to bridge them.

Essentially, although it sounds reasonable to apply saliency detection to WOD, the way to bridge these two tasks is not trivial. The main problem is that saliency detection is formulated as category-free models which only distinguish attractive regions from the image background while irrelevant to the concrete object category. Thus, as shown in Fig. 1, in the images only containing one category of objects (considered as "easy" images), the objects can be captured by saliency detection methods easily and associated with the corresponding image label properly. Whereas in the images weakly labelled as containing multiple categories of objects (considered as "hard" images), objects in all of categories will have the probabilities to attract the human attention and their corresponding locations are also hard for saliency

---

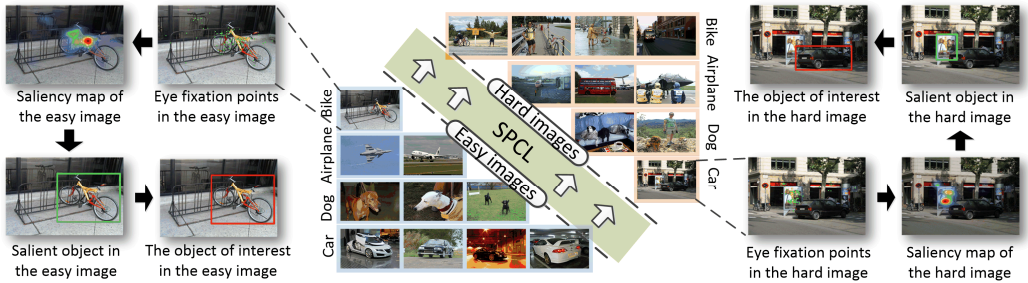

Figure 1: This figure illustrates the main idea of this paper. As can be seen, in the training images with weak labels, some of them (in blue frame) are labelled only containing one object category, which are considered as "easy" images for the saliency detection methods. While others (in pink frame) labelled as containing multiple object categories are considered as "hard" images. Due to the category free property of saliency detection, the objects in "easy" images have larger confidence to be extracted correctly by the saliency detection methods, whereas the objects in "hard" images cannot be extracted successfully. To this end, we develop a novel self-paced curriculum learning paradigm to guide the learner to gradually achieve the faithful knowledge of the multiple object categories from easy (confident) to hard (ambiguous).

models to identify, which largely increases the ambiguity when considering to apply the obtained salient detection results to initializing the training samples for WOD. Thus, it is unreliable to directly apply saliency detection to WOD.

To alleviate this problem, we propose to bridge saliency detection to WOD via a self-paced curriculum learning (SPCL) regime. SPCL was proposed in [Jiang *et al.*, 2015] as a general learning framework including both the curriculum learning (CL) and self-paced learning (SPL) components. To the best of our knowledge, both of these two learning components are critical in successfully bridging saliency detection to WOD, whereas none of the existing literature has explored them before. Specifically, CL was proposed by [Bengio *et al.*, 2009], which is usually learned based on the learning priorities derived by predetermined heuristics for particular problems. SPL was proposed by [Kumar *et al.*, 2010], where the learning pace is dynamically generated by the learner itself, according to which the learner has already learned from the data. Thus, the CL and SPL components in SPCL can be correspondingly used to solve the training sample initialization and object detector updating problems in the proposed saliency-guided WOD. To implement SPCL for our task, we first design a task-specific curriculum to assign the "easy" images with larger priority than the "hard" images during the learning procedure, which indicates that only the salient object hypotheses in the "easy" images are selected as the initial training samples, while the object hypotheses in the "hard" images will be gradually involved in the subsequent learning iterations. To guide the learner to gradually learn faithful knowledge of multi-class objects from the "easy" (high-confidence) images to the "hard" (high-ambiguity) ones, a novel self-paced learning regularizer is proposed to enforce the learner to select confident and diverse training hypotheses in each iteration and learn the object detectors of multiple categories simultaneously. Finally, the proposed SPCL regime can fit well to solve the problems in this paper.

Compared with the SPCL model in [Jiang *et al.*, 2015], the learning regime proposed in this paper mainly has three differences: 1) We design a task-specific learning curriculum for bridging saliency detection and WOD effectively. 2) We introduce an additional term, the sample diversity term, in the self-paced regularizer to prevent the selected training hypotheses from drifting to a small collection of training images. 3) Considering the latent relationship among the multiple categories of co-occurring objects, we further generalize the SPL regime into multi-class formulation, which facilitates the learning system to penalize indiscriminative object hypotheses predicted as belonging to multiple object categories at the same time.

To sum up, there are three-fold contributions in this paper:

- We comprehensively analyze the prospect and challenges in the idea of bridging saliency detection to WOD and propose an effective way to alleviate the problem, which achieves the state-of-the-art detection performance under the weak supervision.
- We establish a novel SPCL regime containing both the task-specific learning curriculum and the data-driven self-learning pace. The regime is well formulated as a concise optimization model.
- We incorporate SPCL with a sample diversity term and further generalize it to work in multi-class scenario.

## 2 Related Works

**Saliency detection:** Most saliency detection methods highlight the attractive image regions by exploring some bottom-up cues. As one frequently explored cue, local contrast [Itti *et al.*, 1998; Klein and Frintrop, 2011] is usually used in the saliency detection models to highlight the image regions appearing differently with their spatially neighbor regions. Another widely used bottom-up cue is the global contrast. Being different from local contrast, global contrast [Achanta *et al.*, 2009; Cheng *et al.*, 2015] is used to discover image regions which are unique in the entire image context. More recently, background prior becomes another important cue for saliency detection. This kind of methods, e.g., [Han *et al.*, 2015], assume that regions near image boundaries are probably backgrounds and detect salient regions as calculating the contrast to these image boundary regions. As can be seen, the bottom-up cues explored by saliency detection models are highly potential to provide helpful priors to the object localizations in each image.

**Algorithm 1:** SPCL for bridging saliency detection to WOD.
    **Input:** Training images with weak labels;
    **Output:** Object detection results in test images.
**1:** Generate object hypotheses and extract the features;
**2:** Initialize **y**, **v** based on salieny detection results according to the learning curriculum; Input parameters $\{\lambda_c, \gamma_c\}_{c=1}^{C}$;
**3: while** *not converge* **do**
**4:**     Update {**W**, **b**} via one-vs-all weighted SVM;
        Update **y** via **Algorithm 2**;
        Update **v** via **Algorithm 3**;
        Renew the pace parameter $\lambda_c$, $\gamma_c$;
**5: end while**
**6:** Detect objects in test images by using the obtained object detectors {**W**,**b**};
**7: return** Object detection results.

**Weakly-supervised object detection:** Two key issues in WOD are 1) predict the potential object localizations and 2) learn the object detectors. Some early WOD methods [Song *et al.*, 2014; Wang *et al.*, 2014] held the view that a better initial estimation of the object localizations is critical to this task as they can largely impact the subsequent learning process. Thus, they explored different ways, e.g. intra-class similarity [Siva and Xiang, 2011], inter-class variance [Siva *et al.*, 2012], and distance mapping relation [Shi *et al.*, 2012], to initialize the training object hypotheses. Later on, some recent WOD methods started to pay more attention to the optimization procedure designed for better training object detectors under the weak supervision. For example, [Bilen *et al.*, 2014; Song *et al.*, 2014] proposed to smooth the object formulation to better obtain the optimal solutions. [Bilen *et al.*, 2015] proposed to incorporate convex clustering in the learning procedure, which enforces the local similarity of the selected hypotheses during optimization.

Essentially, both of the above mentioned problems are critical in WOD task. To this end, this paper proposes a novel SPCL model which explicitly encode both the former problem (with the designed curriculum) and the later (with the self-paced regularizer) into a unified formulation and handle both problems in a theoretically sound manner.

**Self-paced (curriculum) learning:** Inspired by the learning process of humans/animals, the theory of self-paced (or curriculum) learning [Bengio *et al.*, 2009; Kumar *et al.*, 2010] is proposed lately. The idea is to learn the model iteratively from easy to complex samples in a self-paced fashion. By virtue of its generality, the SPL theory has been widely applied to various tasks, such as multi-view clustering [Xu *et al.*, 2015], multi-label propagation [Gong *et al.*, 2016], multimedia event detection [Jiang *et al.*, 2014; Jiang *et al.*, 2014], and co-saliency detection [Zhang *et al.*, 2015]. More recently, [Jiang *et al.*, 2015] introduced the pre-defined learning curriculum to the conventional self-paced learning regime which can take into account both the helpful prior knowledge known before training and the self-learning progress during training. Inspired by this work, we design a task-specific learning curriculum and construct a unified SPCL model specifically for both the saliency detection and WOD tasks, through which both can be naturally related.

## 3 The Proposed Approach

### 3.1 Algorithm Overview

Given $K$ training images with weak labels consisting of $C$ categories, we first extract the bounding box object hypotheses and their corresponding feature representations from each image. Denote the features of each hypothesis in the $k^{\text{th}}$ image as $\{\mathbf{x}_i^{(k)}\}_{i=1}^{n_k}$, $k \in 1, \cdots, K$, where $n_k$ is the hypothesis number in the $k^{\text{th}}$ image. $\mathbf{y} = \{y_{i,c}^{(k)}\}_{i=1,c=1,k=1}^{n_k, \ C, \ K}$ and $\mathbf{v} = \{v_{i,c}^{(k)}\}_{i=1,c=1,k=1}^{n_k, \ C, \ K}$, where $v_{i,c}^{(k)} \in [0,1]$, indicating the labels and the real-valued importance weights of each hypothesis, respectively, which are unknown at the beginning and will be optimized during the proposed learning regime. The aim of the proposed approach is to learn the object detectors {**W**,**b**}, where $\mathbf{W} = \{\mathbf{w}_c\}_{c=1}^{C}$, $\mathbf{b} = \{b_c\}_{c=1}^{C}$, of $C$ object categories from the weakly-labeled training images, and then use them to detect objects in the test images. Specifically, we first design a simple yet effective curriculum to select the salient hypotheses in "easy" images as initialization. Then, the object detectors are trained and updated gradually under the guidance of the proposed self-paced learning strategy. Finally, the obtained object detectors are used to detect the corresponding objects in the test images. The overall algorithm is shown in Algorithm 1.

### 3.2 Problem Formulation

Given object hypotheses from the training images, we propose a simple yet effective curriculum to initialize the learning procedure. Specifically, we first obtain the "easy" images based on the number of weak labels of each image, *i.e.*, images weakly labelled as only containing one object category are considered as "easy". Then, for each "easy" image, we adopt an unsupervised saliency detection method, *i.e.*, RC [Cheng *et al.*, 2015] in this paper due to its efficiency, to generate the corresponding saliency estimation. Finally, the important weights **v** are initialized as the intersection-over-union (IOU) score between each hypothesis and the salient region. The hypotheses with weights larger than 0 are selected as the initial training hypotheses and their labels in **y** are set according to the label of the images containing them.

Afterwards, in order to gradually adapt the learner from the "easy" domain to the "hard" domain and finally capture the faithful knowledge of the objects of interest, a novel self-paced learning regularizer is proposed as follows:

$\min_{\mathbf{W},\mathbf{b},\mathbf{y},\mathbf{v}} E(\mathbf{W},\mathbf{b},\mathbf{y},\mathbf{v}) =$

$\sum_{c=1}^{C} \left( \sum_{k=1}^{K} \sum_{i=1}^{n_k} v_{i,c}^{(k)} \ell \left( y_{i,c}^{(k)}, g\left(\mathbf{x}_i^{(k)}; \mathbf{w}_c, b_c\right) \right) + f_c(\mathbf{v}^{(c)}; \lambda_c, \gamma_c) \right)$

$s.t., y_{i,c}^{(k)} \in \{-1,1\}; \ k=1,\cdots,K; i=1,\cdots,n_k; c=1,\cdots,C$

$\sum_{c=1}^{C} \left| y_{i,c}^{(k)} + 1 \right| \leq 2; \ k=1,\cdots,K; i=1,\cdots,n_k$     (1)

$\sum_{i=1}^{n_k} \left| y_{i,c*}^{(k)} + 1 \right| \geq 2$; if the $k^{\text{th}}$ image is labeled as class $c^*$, where $\mathbf{w}_c, b_c$ represent the SVM parameters for the $c^{\text{th}}$ sub-classification problem. The loss function

$$\ell\left(y_{i,c}^{(k)}, g\left(\mathbf{x}_i^{(k)}; \mathbf{w}_c, b_c\right)\right) = \left(1 - y_{i,c}^{(k)}\left(\mathbf{w}_c^T \mathbf{x}_i^{(k)} + b_c\right)\right)_+ \quad (2)$$

is the hinge loss of the hypothesis $\mathbf{x}_i^{(k)}$ in the $c^{\text{th}}$ sub-classification problem. Three constraints are imposed on labels $\mathbf{y}$. The first one, i.e. $y_{i,c}^{(k)} \in \{-1,1\}$, constrains the label should be binary in each sub-classification problem. The second one, i.e. $\sum_{c=1}^{C} \left| y_{i,c}^{(k)} + 1 \right| \leq 2$, enforces that each hypothesis should belong to only one object category, or no class, i.e., the background category. This constraint inherently penalizes the indiscriminative object hypotheses, i.e. the hypotheses predicted to belong to multiple object categories, when calculating their importance weight in (2). The third one, i.e. $\sum_{i=1}^{n_k} \left| y_{i,c*}^{(k)} + 1 \right| \geq 2$, means that for all object hypotheses located in the $k^{\text{th}}$ image, at least one should belong to the class which the image has been weakly annotated. This will make the learned result finely comply with the prior knowledge.

In the proposed SPCL regime, the self-paced capability is followed by the involvement of the SPL regularizer $f_c(\mathbf{v}^{(c)}; \lambda_c, \gamma_c)$ with the following form:

$$f_c(\mathbf{v}^{(c)}; \lambda_c, \gamma_c) = -\lambda_c \sum_{k=1}^{K} \sum_{i=1}^{n_k} v_{i,c}^{(k)} - \gamma_c \sum_{k=1}^{K} \sqrt{\sum_{i=1}^{n_k} v_{i,c}^{(k)}}, \quad (3)$$

where $\lambda_c, \gamma_c$ are the class-specific parameters imposed on the easiness term and the diversity term, respectively. $\mathbf{v}^{(c)}$ indicates the important weights of each hypothesis to the $c^{\text{th}}$ object category.

The negative $l_1$-norm term is inherited from the conventional SPL [Kumar *et al.*, 2010], which favors selecting easy over complex hypotheses. If we omit the diversity term, i.e. let $\gamma_c = 0$, the regularizer degenerates to the traditional hard SPL function proposed in [Kumar *et al.*, 2010], which conducts either 1 or 0 (i.e. selected in training or not) for the weight $v_{i,c}^{(k)}$ imposed on hypothesis $\mathbf{x}_i^{(k)}$, by judging whether its loss value is smaller than the pace parameter $\lambda_c$ or not. That is, a sample with smaller loss is taken as an easy sample and thus should be learned preferentially and vice versa.

Another regularization term favors selecting diverse hypotheses residing in more images. This can be easily understood by seeing that its negative leads to the group-wise sparse representation of $\mathbf{v}$. Contrariwise, this diversity term should have a counter-effect to group-wise sparsity. That is, minimizing this diversity term tends to disperse non-zero elements of $\mathbf{v}$ over more images, and thus favors selecting more diverse hypotheses. Consequently, this anti-group-sparsity representation is expected to realize the desired diversity. Different from the commonly utilized $l_{2,1}$ norm [Jiang *et al.*, 2014], our utilized group-sparsity term is concave, leading to the convexity of its negative. This on one side simplifies the designation of the solving strategy, and on the other hand well fits the previous axiomatic definition for the SPL regularizer [Jiang *et al.*, 2014].

### 3.3 Optimization Method

The solution of (1) can be approximately attained by alternatively optimizing the involved parameters $\{\mathbf{W}, \mathbf{b}\}$, $\mathbf{y}$ and $\mathbf{v}$ as described in Algorithm 1. The optimization mainly contains following steps:

**Object detectors updating:** Optimize object detector parameters $\{\mathbf{W}, \mathbf{b}\}$ via one-vs-all SVM under fixed $\mathbf{y}$ and $\mathbf{v}$. In this case, (1) degenerates to the following form:

$$\min_{\mathbf{W},\mathbf{b}} \sum_{c=1}^{C} \left( \sum_{k=1}^{K} \sum_{i=1}^{n_k} v_{i,c}^{(k)} \ell\left(y_{i,c}^{(k)}, g\left(\mathbf{x}_i^{(k)}; \mathbf{w}_c, b_c\right)\right) \right), \quad (4)$$

which can be equivalently reformulated as solving the following sub-optimization problems for each $c=1,2,\cdots,C$:

$$\min_{\mathbf{w}_c, b_c} \sum_{k=1}^{K} \sum_{i=1}^{n_k} v_{i,c}^{(k)} \ell\left(y_{i,c}^{(k)}, g\left(\mathbf{x}_i^{(k)}; \mathbf{w}_c, b_c\right)\right). \quad (5)$$

This is a standard one-vs-all (weighted) SVM model [Yang *et al.*, 2007].

**Hypotheses labelling:** Optimize $\mathbf{y}$ under fixed $\{\mathbf{W}, \mathbf{b}\}$ and $\mathbf{v}$: The goal of this step is to learn the pseudo-labels of training hypotheses from the current object detectors. The model in this case can be reformulated as:

$$\min_{\mathbf{y}} \sum_{c=1}^{C} \left( \sum_{k=1}^{K} \sum_{i=1}^{n_k} v_{i,c}^{(k)} \ell\left(y_{i,c}^{(k)}, g\left(\mathbf{x}_i^{(k)}; \mathbf{w}_c, b_c\right)\right) \right)$$

$s.t., y_{i,c}^{(k)} \in \{-1,1\}; k = 1,\cdots,K; i = 1,\cdots,n_k; c = 1,\cdots,C$

$\sum_{c=1}^{C} \left| y_{i,c}^{(k)} + 1 \right| \leq 2; k = 1,\cdots,K; i = 1,\cdots,n_k \quad (6)$

$\sum_{i=1}^{n_k} \left| y_{i,c*}^{(k)} + 1 \right| \geq 2$; if the $k^{\text{th}}$ image is labelled as class $c^*$.

This problem can be equivalently decomposed into sub-problems with respect to each $k = 1,\cdots,K$, i.e. for each image, where $c^*$ is the weak labels of the $k^{\text{th}}$ image:

$$\min_{\mathbf{y}^{(k)}} \sum_{c=1}^{C} \sum_{i=1}^{n_k} v_{i,c}^{(k)} \ell\left(y_{i,c}^{(k)}, g\left(\mathbf{x}_i^{(k)}; \mathbf{w}_c, b_c\right)\right)$$

$s.t., y_{i,c}^{(k)} \in \{-1,1\}; i = 1,\cdots,n_k; c = 1,\cdots,C$

$\sum_{c=1}^{C} \left| y_{i,c}^{(k)} + 1 \right| \leq 2; i = 1,\cdots,n_k \quad (7)$

$\sum_{i=1}^{n_k} \left| y_{i,c*}^{(k)} + 1 \right| \geq 2$; if the $k^{\text{th}}$ image is labelled as class $c^*$,

where $\mathbf{y}^{(k)} = \{y_{i,c}^{(k)}\}_{i=1,c=1}^{n_k, C}$ indicates the labels of the hypotheses in the $k^{\text{th}}$ image. Its global optimum can be attained by Algorithm 2, which can be derived from the theorem in [Zhang *et al.*, 2015].

**Hypotheses re-weighting:** Optimize $\mathbf{v}$ under fixed $\{\mathbf{W}, \mathbf{b}\}$ and $\mathbf{y}$: After updating the pseudo-labels, we aim to renew the weights on all hypotheses to reflect their different importance to learning of the current decision surface. In this case, (1) degenerates to the following form:

$$\min_{\mathbf{v}} \mathbf{E}(\mathbf{v}) = \sum_{c=1}^{C} \left( \sum_{k=1}^{K} \sum_{i=1}^{n_k} v_{i,c}^{(k)} \ell\left(y_{i,c}^{(k)}, g\left(\mathbf{x}_i^{(k)}; \mathbf{w}_c, b_c\right)\right) + f_c(\mathbf{v}^{(c)}; \lambda_c, \gamma_c) \right), \quad (8)$$

which is equivalent to independently solving the following sub-optimization problem for each $k = 1,\cdots,K$ and $c = 1,\cdots,C$ via:

$$\min_{v_{i,c}^{(k)} \in [0,1], i=1,\cdots,n_k} \sum_{i=1}^{n_k} v_{i,c}^{(k)} \ell\left(y_{i,c}^{(k)}, g\left(\mathbf{x}_i^{(k)}; \mathbf{w}_c, b_c\right)\right)$$

$$+ \sum_{i=1}^{n_k} v_{i,c}^{(k)} - \gamma_c \sqrt{\sum_{i=1}^{n_k} v_{i,c}^{(k)}}. \quad (9)$$

We can easily simplify the above optimization problem as:

**Algorithm 2:** Algorithm of Optimizing $\mathbf{y}^{(k)}$

**Input:** Hypotheses $\{\mathbf{x}_i^{(k)}\}_{i=1}^{n_k}$, object detector $\{\mathbf{W},\mathbf{b}\}$, importance weight $\{v_{i,c}^{(k)}\}$, the weak label $c^*$ for this image;
**Output:** All Pseudo-labels in $\mathbf{y}^{(k)}$.

1: **for** $i = 1, \cdots, n_k$
2:    **if** $\mathbf{w}_c^T \mathbf{x}_i^{(k)} + b_c < 0$ for all $c = 1, \cdots, C$
3:    **then** $\bar{y}_{i,c}^{(k)} = -1$ for all $c = 1, \cdots, C$
4:    **otherwise** $\bar{y}_{i,c}^{(k)} = \begin{cases} +1, c = \hat{c} \\ -1, c \neq \hat{c} \end{cases}$,
     where $\hat{c} = \operatorname{argmax}_{1 \leq c \leq C} \mathbf{w}_c^T \mathbf{x}_i^{(k)} + b_c$
5: **end for**
6: **if** $\sum_{i=1}^{n_k} \left| \bar{y}_{i,c^*}^{(k)} + 1 \right| < 2$; **then**
7: **for** $i = 1, \cdots, n_k$
8:    **if** $\sum_{c=1}^{C} |y_{i,c}^{(k)} + 1| = 0$
9:    **then** $\Delta_i = v_{i,c^*}^{(k)} \ell\left(1, g(\mathbf{x}_i^{(k)}; \mathbf{w}_{c^*}, b_{c^*})\right)$, case=0;
10:    **otherwise**
     $\Delta_i = v_{i,c^*}^{(k)} \ell\left(1, g(\mathbf{x}_i^{(k)}; \mathbf{w}_{c^*}, b_{c^*})\right) + v_{i,\hat{c}}^{(k)} \ell\left(1, g(\mathbf{x}_i^{(k)}; \mathbf{w}_{\hat{c}}, b_{\hat{c}})\right)$
     where $\hat{c} = \operatorname{argmax}_{1 \leq c \leq C} \mathbf{w}_c^T \mathbf{x}_i^{(k)} + b_c$, case=1;
11: **end for**
12: $i^* = \operatorname{argmin}_{i=1,\cdots,n_k} \Delta_i$
13: **if** case=0, then $y_{i^*,c^*}^{(k)} = 1$;
14: **if** case=1, then $y_{i^*,c^*}^{(k)} = 1$, $y_{i^*,\hat{c}}^{(k)} = -1$;
15: **return** $\mathbf{y}^{(k)}$.

$$\min_{\substack{v_i \in [0,1], \\ i=1,\cdots,n}} \mathbf{E}(\mathbf{w}, b, \mathbf{y}, \mathbf{v}) = \sum_{i=1}^n v_i l_i - \lambda \sum_{i=1}^n v_i - \gamma \sqrt{\sum_{i=1}^n v_i}. \quad (10)$$

This model is convex and according to [Zhang *et al.*, 2015], we can apdopt an effective algorithm, *i.e.*, the Algorithm 3, for extracting the global optimum to it.

## 4 Experimental Results

### 4.1 Experimental Settings

We evaluate our method on the Pascal VOC 2007 dataset [Everingham *et al.*, 2008] which is widely used by the previous works. In our experiments, we follow the previous works [Deselaers *et al.*, 2012; Bilen *et al.*, 2014; Bilen *et al.*, 2015; Shi *et al.*, 2015] to discard any images that only contain object instances marked as "difficult" or "truncated" during the training phase, while all the images in the VOC07-Test are used during the test phase. For fair comparison, we follow the standard VOC procedure [Everingham *et al.*, 2008] and report average precision (AP) on the Pascal VOC 2007 *test* split.

Being consistent with the recently proposed WOD methods [Bilen *et al.*, 2014; Song *et al.*, 2014; Bilen *et al.*, 2015], we apply Selective Search [Uijlings *et al.*, 2013] to generate around 1500 bounding box object hypotheses in each image and adopt the CNN features [Donahue *et al.*, 2013] pre-trained on the ImageNet 2012 to represent each of the extracted object hypotheses. Before training, $\lambda_c$ and $\gamma_c$ in (2) need to be set in advance. As suggested in [Jiang *et al.*, 2014; Jiang *et al.*, 2014], we set $\lambda_c$ according to the number of the selected training hypotheses which is set to be 2% of the total bounding box windows extracted from the images

**Algorithm 3:** Algorithm of Optimizing $\mathbf{v}$.

**Input:** Object hypotheses $\{\mathbf{x}_i\}_{i=1}^n$ with their current labels $\mathbf{y}$, object detector $\{\mathbf{W}, \mathbf{b}\}$, parameters $\lambda$ and $\gamma$;
**Output:** The global solution $\mathbf{v}$ of (10).

1: sort $\mathbf{x}_1, \mathbf{x}_2, \cdots \mathbf{x}_n$ in ascending order of their loss values, *i.e.*, $l_1 \leq l_2 \leq \cdots \leq l_n$; Let $m=0$;
2: **for** $i =1$ **to** $n$ **do**
2:    **if** $l_i < \lambda + \gamma/(2\sqrt{i})$ **then** $v_i = 1$;
3:    **if** $l_i \geq \lambda + \gamma/(2\sqrt{i})$ **then** count the number $m$ of $l_j = l_i$ for $j=i, i+1, \cdots, n$, let $v_i = \cdots = v_{i+m-1} = \left(\left(\gamma/2(l_i - \lambda)\right)^2 - (i-1)\right)/m$ and $v_{i+m} = \cdots = v_n = 0$; **Break;**
4: **end for**
5: **return** $\mathbf{v}$.

weakly labelled as containing the $c^{\text{th}}$ object category. Then, $\gamma_c$ is set to be equal to $\lambda_c$ empirically.

### 4.2 Comparison to the State-of-the-arts

In this section, we evaluate the object detection performance of our framework by comparing it with 6 state-of-the-art WOD approaches which are PR [Bilen *et al.*, 2014], CC [Bilen *et al.*, 2015], MDD [Siva and Xiang, 2011], LLO [Song *et al.*, 2014], VPC [Song *et al.*, 2014], and MfMIL [Cinbis *et al.*, 2014]. For quantitative comparison, we report the evaluation results in terms of the AP score in Fig. 2. As can be seen, the proposed approach obtains the highest score of 29.96 on average. According to our analysis, the proposed approach can obtain significantly better results than MDD and MfMIL mainly due to the better feature representation, stronger saliency prior, and more powerful learning scheme. Compared with PR, CC, LLO, and VPC, the performance gain of the proposed approach mainly comes from the core insight of this paper, *i.e.*, developing property way to bridge saliency detection to WOD, as we used the same feature representation with these methods. More specifically, compared with PR and CC, the performance gain of the proposed approach comes mainly from the idea to bridge saliency detection to WOD because they only adopted weak priors in their initialization. Compared with LLO and VPC, the performance gain of the proposed approach mainly comes from the proposed SPCL regime as these two methods also explored strong prior information for initializing the training hypotheses in their frameworks.

Some examples of the detection results are also shown in Fig. 2, which includes some successful cases, *i.e.*, the examples in the *bus* and *cat* categories, as well as some failure cases, *i.e.*, examples in the *plant* and *chair* categories. The successful cases subjectively demonstrate the effectiveness of the proposed approach. For the failure cases, the main problem is that very limited number of images only contains the objects like *plant* and *chair*, leading to the insufficient training hypotheses in the initialization stage. This problem can be solved by designing more proper learning curriculum for WOD in future works.

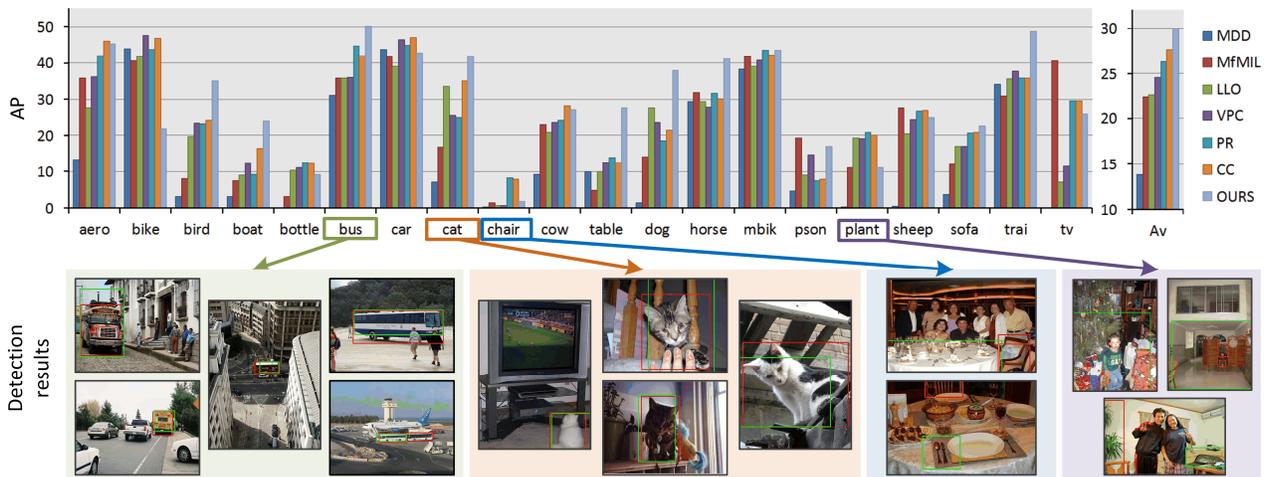

Figure 2: Quantitative and subjective evaluation of the proposed approach. The left two blocks of the detection results are the successful examples, while the right two blocks are the failure examples. The red boxes are the ground-truth annotations. The green ones are the detection results (corresponding to the specified categories) of the proposed weakly-supervised framework.

Table 1: Definitions of the baseline models.

| Name | Definition |
| --- | --- |
| Sal+SVM | Directly train SVM using the salient hypotheses in each image. |
| Sal+SPL | Bridge saliency detection and WOD via a basic SPL regime [Kumar et al., 2010]. |
| Sal+SPCL | Bridge saliency detection and WOD via a basic SPCL regime [Jiang et al., 2015]. |
| LLO | Baseline WOD framework [Song et al., 2014]. |
| LLO+SPCL* | Replace the learning model, i.e., SLSVM, in LLO with the proposed SPCL regime. |

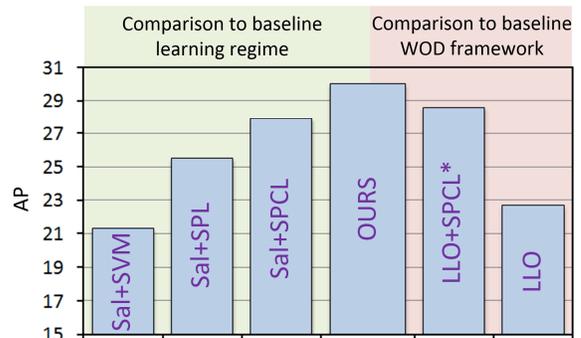

Figure 3: Quantitative comparisons to baseline models.

### 4.3 Model Analysis

To further analyze the proposed framework in this paper, we make more comprehensive evaluations in this section by comparing with five baseline models as described in Table 1. The experimental results are shown in Fig. 3, from which we can see: 1) The performance gap between Sal+SVM and OURS demonstrates the importance to develop proper ways to bridge saliency detection and WOD. 2) The experimental results of Sal+SPL, Sal+SPCL, and OURS demonstrate the better performance of the proposed learning regime as compared with some existing self-paced (curriculum) learning regimes. 3) The performance gap between OURS and LLO+SPCL* demonstrates the saliency prior can provide more helpful information than the prior designed in LLO. 4) The performance gap between LLO+SPCL* and LLO indicates the better capability of the proposed learning regime as compared with the learning model in one state-of-the-art WOD framework. According to the above analysis, the key insight of this paper, i.e., developing powerful learning regime, i.e., the proposed SPCL, can better bridge saliency detection to WOD and help the learner to capture the faithful knowledge of the object categories under weak supervision, has been demonstrated comprehensively.

### 5 Conclusion

In this paper, in order to address the challenging WOD problem, we proposed an effective framework to bridge saliency detection to the investigated task based on a novel SPCL regime. The insight of this paper is that by developing powerful learning regime which contains both the task-specific learning curriculum and the data-driven self-learning pace, saliency detection technique can be better leveraged to provide beneficial information for helping the learner to capture the faithful knowledge of the object categories under weak supervision. Experiments including comparisons to other state-of-the-arts and comprehensive analysis of the proposed framework on the benchmark dataset have demonstrated the effectiveness of our approach. For the future work, inspired by [Long et al., 2014], we plane to enable the proposed method to transfer the knowledge that has be captured to new concepts via novel regularizers.

**Acknowledgments:** This work was supported in part by the National Science Foundation of China under Grants 61522207 and 61473231, the Doctorate Foundation, and the Excellent Doctorate Foundation of Northwestern Polytechnical University.